\documentclass{article}
\usepackage{arxiv}
\usepackage{multirow}
\usepackage{subcaption}
\usepackage{diagbox}
\usepackage{bbding}
\usepackage{makecell}
\usepackage{graphicx}

\usepackage{hyperref}

\usepackage{amsmath}
\usepackage[utf8]{inputenc} 
\usepackage[T1]{fontenc}    
\usepackage{url}            
\usepackage{booktabs}       
\usepackage{amsfonts}       
\usepackage{nicefrac}       
\usepackage{microtype}      
\usepackage{cleveref}       
\usepackage{lipsum}         
\usepackage{natbib}
\usepackage{doi}

\usepackage{xcolor}
\hypersetup{
    colorlinks,
    linkcolor={red!50!black},
    citecolor={blue!50!black},
    urlcolor={blue!80!black}
}

\AtBeginDocument{%
  }

\begin{document}


\title{From Stars to Insights: Exploration and Implementation of Unified Sentiment Analysis with Distant Supervision}
\renewcommand{\shorttitle}{Unified Sentiment Analysis with Distant Supervision}



\author{
    \href{https://orcid.org/0000-0001-9323-2932}{\includegraphics[scale=0.06]{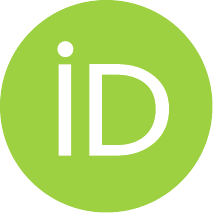}\hspace{1mm}Wenchang Li} \\
    Peking University\\
    Beijing, China\\
    \texttt{wli@stu.pku.edu.cn}\\
    \And
    \href{http://orcid.org/0000-0003-0848-4786}{\includegraphics[scale=0.06]{orcid.pdf}\hspace{1mm}John P. Lalor},
    \href{http://orcid.org/0000-0001-5509-4161}{\includegraphics[scale=0.06]{orcid.pdf}\hspace{1mm}Yixing Chen}, 
    \href{http://orcid.org/0000-0002-6228-8017}{\includegraphics[scale=0.06]{orcid.pdf}\hspace{1mm}Vamsi K. Kanuri} \\
    University of Notre Dame\\
  Notre Dame, Indiana, USA\\
\texttt{john.lalor@nd.edu}, \texttt{ychen43@nd.edu}, \texttt{vkanuri@nd.edu}\\
}


\date{}

\maketitle

\begin{abstract}

Sentiment analysis is integral to understanding the voice of the customer and
informing businesses' strategic decisions. Conventional sentiment analysis
involves three separate tasks: aspect-category detection, aspect-category
sentiment analysis, and rating prediction. However, independently
tackling these tasks can overlook their interdependencies and often requires
expensive, fine-grained annotations. This paper introduces unified sentiment
analysis, a novel learning paradigm that integrates the three aforementioned tasks into
a coherent framework. To achieve this, we propose the Distantly Supervised
Pyramid Network (DSPN), which employs a pyramid structure to capture sentiment
at word, aspect, and document levels in a hierarchical manner. Evaluations on
multi-aspect review datasets in English and Chinese show that DSPN, using only
star rating labels for supervision, demonstrates significant efficiency
advantages while performing comparably well to a variety of benchmark models.
Additionally, DSPN's pyramid structure enables the interpretability of its
outputs. Our findings validate DSPN's effectiveness and efficiency, establishing
a robust, resource-efficient, unified framework for sentiment analysis. 

\footnote{Published version available at \url{http://dx.doi.org/10.1145/3757747}.}

\end{abstract}





\section{Introduction}
\label{sec:intro}

As markets grow more competitive, businesses are increasingly turning to
customer feedback for richer insights that can help them address customer pain
points, enhance customer satisfaction, and drive growth. Online reviews have
become a vital resource for gathering these insights because they provide
unfiltered feedback from consumers with direct experience of a company's
products and services~\citep{zhang2023uncovering, kang2019helpfulness}, and
capture a wide spectrum of customer experiences, including those involving
product/service features, service quality, price, and brand reputation.

Consequently, researchers have prescribed several frameworks to analyze online
reviews. A bulk of those primarily focus on determining the overall sentiment
within a review~\citep{liu2012survey}. However, by concentrating only on an
aggregate sentiment score, such analyses can overlook more detailed feedback
within reviews. This limitation may obscure actionable insights that could drive
targeted improvements in customer service or product design.

\begin{figure}[t]
	\centering 
	\includegraphics[width=0.75\columnwidth]{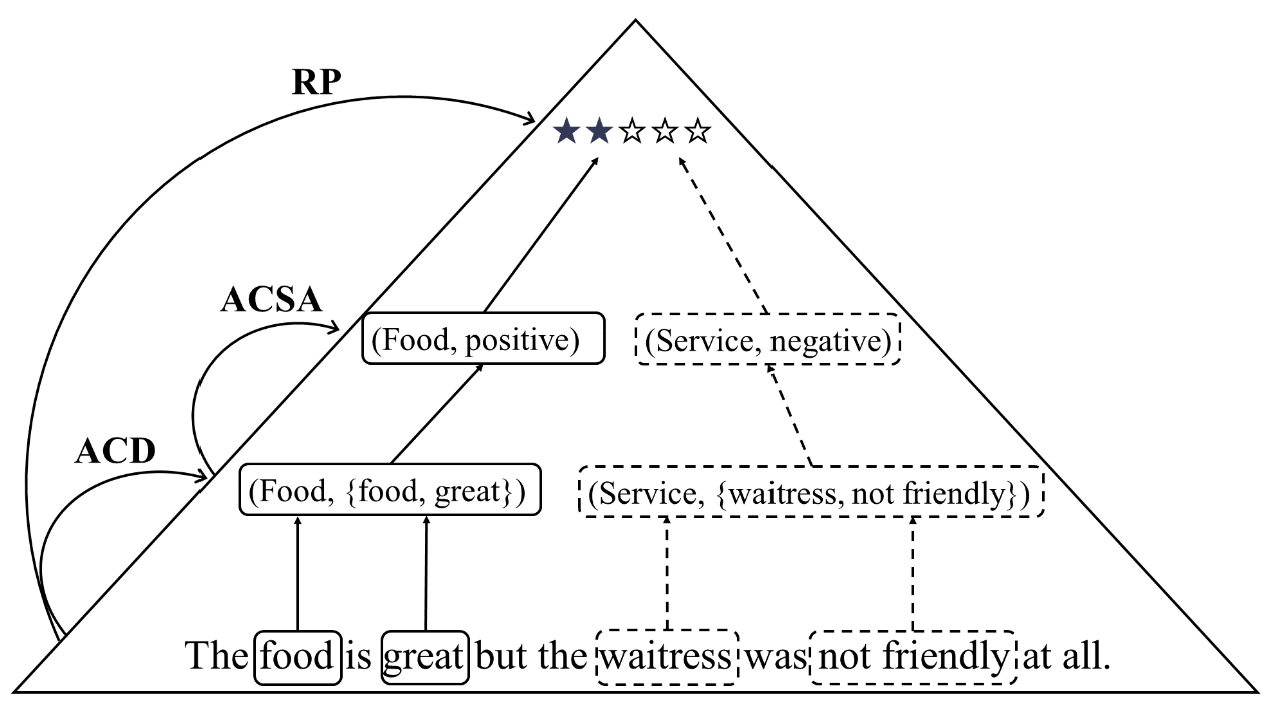}
	\caption{An Overview of Unified Sentiment Analysis. While aspect-category detection (ACD), aspect-category sentiment analysis (ACSA), and rating prediction (RP) can be performed individually, leveraging the implicit pyramid structure
	of reviews enables efficient completion of all three tasks using only RP
	labels.} 
	\label{FigOverview} 
\end{figure}

Consider, for example, the review shown in Figure~\ref{FigOverview}:
``\textit{The food is great, but the waitress was not friendly at all}.''
Traditional sentiment analysis might interpret this review as portraying a
neutral sentiment overall. In contrast, a more granular approach reveals that,
while the customer enjoyed the food, they were dissatisfied with the service.
This type of fine-grained analysis allows businesses to capture richer and more
nuanced sentiments---such as a positive response to food and a negative response
to service---that can lead to targeted improvements, such as enhancing customer
service while maintaining food quality.

To obtain such nuanced insights from an online review, an analyst would need to
perform three key tasks: identify key aspects within a review (i.e.,
aspect-category detection, or ACD), classify the sentiment associated with each
aspect (i.e., aspect-category sentiment analysis, or ACSA), and predict the
overall sentiment or rating of the review (i.e., rating prediction, or RP). For
example, a model would use ACD to identify aspects such as \textit{Food} and
\textit{Service} in the aforementioned example, apply ACSA to assess their
sentiment (e.g., \textit{Food: Positive}, \textit{Service: Negative}), and use
RP to predict an overall rating (e.g., two stars). 
Scholars in information systems have explored these three types 
of sentiment analysis models.
For instance, ACD models were used to measure text richness~\citep{qiao2023text} 
or to extract reasons for medication nonadherence~\citep{xie2022understanding}.
ACSA models were used to extract consumer opinions and their finer-grained 
interactions~\citep{zhang2023uncovering}, as well as to establish relationships between 
online reviews and product sales~\citep{jiang2021investigating}.
RP has also been widely studied in information systems literature~\citep[e.g.,][]{bauman2022know,wang2025predicting,zimbra2018twitter}.

A key limitation of prior studies is that they treat the ACD, ACSA, and RP tasks independently.\footnote{In some cases, ACD and ACSA are
combined~\citep[e.g.,][]{schmitt2018joint, liu2021solving}. In particular, RP is rarely
integrated into aspect-level analyses~\citep{chebolu2023review}.} Therefore, 
training models for these tasks requires
large, annotated datasets for each task, making dataset construction 
a costly and time-intensive process.
Such annotation is typically done independently, limiting the ability to model 
the relationship between 
aspect-level sentiment and overall rating.
Producing aspect-based labels is particularly labor-intensive, as each review
must be manually tagged for sentiments tied to specific attributes.
Additionally, prior research seldom incorporates the hierarchical structure in
sentiment data, which ranges from word-level expressions (e.g., ``delicious'')
to aspect-level (e.g., ``food'') and overall sentiment (e.g., ``five stars'').
Not accounting for such hierarchical structure within unstructured text data can
result in efficiency loss and poor predictive accuracy \citep[e.g.,][for ACD and
ACSA]{chen-qian-2020-relation}.

To address these challenges, we propose an integrated approach to sentiment
analysis, which we refer to as \textit{unified sentiment analysis}. Unified sentiment analysis is grounded
in the idea that a review's overall rating reflects the cumulative sentiments
associated with various aspects and that aspect-level sentiments are informed by
word-level expressions within the review~\citep{bu2021asap}. 
To model unified sentiment analysis, researchers can model each task independently and combine 
predictions using a pipeline approach. 
However, by merging tasks and modeling them in an end-to-end manner, 
models can leverage interactions between the levels for improved predictive
performance and interpretability.

In this work, we propose the Distantly Supervised Pyramid Network (DSPN), 
a model for learning unified sentiment analysis in an end-to-end fashion. 
Notably, DSPN
can unify the ACD, ACSA, and RP tasks by using \textit{only} review-level star
ratings for training.  DSPN employs a pyramid structure to capture sentiment at
each level—word, aspect, and review. Specifically, the model begins by
identifying word-level sentiments, aggregates them to determine aspect-level
sentiments, and then synthesizes these to predict the review's overall rating.
For instance, in our prior example, recall that the reviewer stated that the
food was good, but the service quality was unacceptable. Considering the two
aspects (i.e., food and service), the reviewer gave the restaurant a two-star
rating (Figure \ref{FigOverview}). DSPN holistically captures these nuances by
acknowledging that the aspect-level preferences drive the overall review of two
stars (out of a possible five). By relying solely on readily available review
star ratings for training~\citep{li2020understanding}, DSPN eliminates the need
for costly, manually annotated aspect-level labels, making it an efficient and
cost-effective tool for comprehensive sentiment analysis. Our empirical results
show that this efficient approach can improve performance for the RP task,
indicating that the pyramid structure can both make predictions for unlabeled
tasks while maintaining performance on the main labeled task.
DSPN is also built in a modular fashion, so that improvements in one of the unified sentiment analysis
tasks can be incorporated seamlessly into DSPN.
Our model also offers a novel extension to 
an existing ACD method ~\citep{he2017unsupervised}. DSPN not only improves ACD performance but also improves ACSA and PR performance.

We aim to make three contributions to the information systems literature. First, we introduce unified sentiment analysis
as a novel task formulation that integrates three important but traditionally separate sentiment analysis tasks: ACD, ACSA, and RP. We formally give definitions and notations of
unified sentiment analysis and its constituent parts (see Section \ref{sec:unisa}) that future
research can follow. Second, we propose DSPN as a novel model for the unified sentiment analysis task. DSPN captures the pyramid sentiment
structure effectively and shows significant efficiency gains with only RP labels
as training signal. Third, we validate DSPN through experimental results on English and
Chinese multi-aspect datasets and demonstrate its effectiveness and
efficiency.\footnote{Our code is available at
\url{https://github.com/nd-ball/DSPN}.}

From a managerial perspective, our unified framework offers practical advantages for extracting actionable insights from unstructured data such as customer reviews, employee feedback, or social media posts. Traditionally, organizations have relied on separate models to analyze fine-grained sentiment (e.g., on specific product features) and overall sentiment (e.g., satisfaction or brand perception), leading to fragmented insights and missed connections between various types of concerns. By integrating these layers into a single, interpretable model, our proposed approach enables managers to understand how specific experiences or issues contribute to broader sentiment outcomes. This makes it possible to directly link aspect-level feedback to higher-level performance indices such as ratings, retention, or loyalty. Furthermore, DSPN’s ability to operate effectively with only document-level labels reduces reliance on costly manual annotation, making it especially useful in data-rich but label-scarce environments.

\section{Related Work}
\label{sec:relatedwork}

Comprehensive literature reviews suggest that sentiment analysis 
extensively studied in the literature  
includes three key tasks: aspect-category detection (ACD), aspect-category sentiment analysis
(ACSA), and rating prediction (RP)~\citep{liu2012survey,schouten2015survey}. Table \ref{tb1} lists the most
relevant studies, which serve as benchmarks in our experiments later. Below, we
briefly review these studies within the context of our research.


\begin{table}[h!]
\footnotesize
\centering
\begin{tabular}{cccccc}
\toprule
\multirow{2}{*}{\textbf{Reference}} & \multicolumn{3}{c}{\textbf{Supported
Tasks}} & \multirow{2}{*}{\textbf{Supervision}} & \multirow{2}{*}{\textbf{Model
Architecture}} \\
\cmidrule{2-4}
~ & \textbf{ACD} & \textbf{ACSA} & \textbf{RP} & ~ & ~ \\
\midrule
    \citet{he2017unsupervised} & \Checkmark &   &  & Unsupervised & AutoEncoder
    \\
    \citet{xue2018aspect} &   & \Checkmark & & Supervised & Gated Convolutional
    Networks \\
    \citet{sun2019fine} &   &   & \Checkmark & Supervised & BERT \\
    \citet{schmitt2018joint} & \Checkmark & \Checkmark & & Supervised & LSTM,
    CNN \\
    \citet{li2020multi} & \Checkmark & \Checkmark & & Supervised & BERT \\
    \citet{liu2021solving} & \Checkmark & \Checkmark & & Supervised & BERT, BART
    \\
    \midrule
    This paper & \Checkmark & \Checkmark & \Checkmark & \makecell{Unsupervised
    (Module 1) \\ Distantly Supervised (Module 2)} & BERT \\
\bottomrule
\end{tabular}
\caption{\label{tb1} Comparison with representative approaches on supported
tasks, supervision, and model architecture. ACD, ACSA, and RP refers to aspect-category detection, aspect-category sentiment analysis, and rating prediction respectively.}
\end{table}

\subsection{Aspect-Category Detection}

ACD methods can be categorized into rule-based, supervised, or unsupervised methods.
Rule-based approaches~\cite[e.g.,][]{hai2011implicit, schouten2014commit} rely
on manually defined rules and domain knowledge. Supervised
methods~\cite[e.g.,][]{toh2016nlangp, xue2017mtna} require reviews annotated
with predefined aspect categories. Unsupervised
models~\cite[e.g.,][]{titov2008modeling, brody2010unsupervised, zhao2010jointly}
extract aspects by identifying word co-occurrence patterns. The attention-based aspect extraction (ABAE)
model~\citep{he2017unsupervised}, which employs an autoencoder-style network, is
a foundational unsupervised approach and forms the basis of our Module 1.
Recently, \citet{tulkens2020embarrassingly} introduced a model that combines part-of-speech 
tagging and word embeddings with a contrastive attention mechanism,
outperforming more complex methods. Extending prior work, we propose a novel
aspect-attention mechanism that integrates ACD outputs into the ACSA task.


\subsection{Aspect-Category Sentiment Analysis}
Currently, most ACSA methods are supervised~\citep{schouten2015survey,li2020multi,liu2021solving} and require
costly, time-intensive aspect-level data annotation. Unsupervised LDA-based ACSA
models~\cite[e.g.,][]{zhao2010jointly, xu2012towards, garcia2018w2vlda} often
depend on external resources such as parts-of-speech tagging and sentiment word
lexicons. However, they can suffer from a topic-resemblance problem~\citep{huang2020weakly}.
To address this issue, \citet{huang2020weakly} proposed JASen, a weakly-supervised
approach that learns a joint aspect-sentiment topic embedding. JASen 
can only model documents with a single annotated aspect (either positive or negative), 
effectively reducing the task to RP and limiting its generalizability. 
Recently, \citet{kamila2022ax} introduced AX-MABSA, an
extremely weakly supervised ACSA model that achieves strong performance without
any labeled data.
The current paper advances prior research by proposing a distantly supervised
pyramid network that efficiently performs the ACSA task using only star rating
labels.


\subsection{Rating Prediction}
RP, or review-level sentiment analysis, is used to predict user opinions from
large review datasets. Typically
modeled as a multi-class classification task, RP has been
well-studied~\cite[e.g.,][]{ganu2009beyond, li2011incorporating, liu2012survey,
chen2018neural, bauman2022know}. Recent advances focus on improving interpretability and
prediction accuracy. For example, the CEER framework~\citep{zhou2024enhancing}
enhances explanation generation in recommender systems by utilizing macro
concepts from user-item reviews. \citet{SEJWAL2022118307} introduced RecTE,
which uses topic embeddings to predict ratings and address cold-start issues.
These approaches show the benefits of incorporating textual data into RP models
to boost performance and user trust. However, the pyramid sentiment structure
has not yet been leveraged in RP models to assist in aspect-level sentiment
learning.

\subsection{Multi-Task Sentiment Analysis}
There has been some work on jointly learning ACSA and RP~\citep{bu2021asap}, using RP
information for ACSA~\citep{yin2017document, li2018document, he2018exploiting},
and using ACSA information for RP~\citep{cheng2018aspect, wu2019arp}. Prior
approaches on document-level multi-aspect sentiment classification predict user
ratings for different product or service aspects~\citep{yin2017document,
li2018document}, often incorporating user information and star ratings to boost
performance, though at the cost of efficiency. Other works~\citep{bu2021asap,
fei2022making} have jointly learned ACD and ACSA; these methods require costly
aspect-level data annotation, hindering efficiency. For instance,
\citet{schmitt2018joint} proposed joint models for ACD and ACSA in an end-to-end
manner. To our knowledge, our work is among the first that
simultaneously models all three tasks with a single task source, RP labels, for
supervision.

To enhance the readability of this paper, we summarize the most frequently used acronyms in Table \ref{tab:glossary}.

\begin{table}[h!]
    \centering
    \small
\begin{tabular}{ll}
    \toprule
    \textbf{Acronym} & \textbf{Expansion}  \\
    \midrule
     DSPN & Distantly Supervised Pyramid Network  \\
     ACD & Aspect-Category Detection \\
     ACSA & Aspect-Category Sentiment Analysis  \\
     RP &  Rating Prediction \\
     ABAE & Attention-Based Aspect Extraction \\
    \bottomrule
\end{tabular}
\caption{\label{tab:glossary} Glossary of acronyms used.}
\end{table}

\section{Unified Sentiment Analysis}
\label{sec:unisa}
Before describing our model, we first define our notation and present the
unifying framework of unified sentiment analysis. We borrow notation from the prior work where
possible and introduce new notation as needed for consistency across
tasks~\citep{pontiki2016semeval}. For clarity and consistency, we provide a
comprehensive description of the notation we use in this article (Table
\ref{tab:variables}).
Our corpus is a collection of \textit{reviews} $\mathbf{R} = \{R_1, R_2, \dots,
R_{\vert \mathbf{R} \vert} \}$. Each review $R_i$ consists of a sequence of word
tokens (hereafter ``words''): $R_i = \{t_i^{(1)}, t_i^{(2)}, \dots,
t_i^{(n)}\}$.

\begin{table}[h!]
\centering
\scalebox{0.9}{
\renewcommand{\arraystretch}{1.2}
\begin{tabular}{cp{9.5cm}c}
\toprule
\textbf{Variable} & \textbf{Description} & \textbf{Dimension} \\
\midrule
    $d_w$ & Embedding dimension & $\mathbb{R}^{768}$ \\
    $\mathbf{R}$ & Reviews in our dataset & - \\
    $R_i$ & $i$-th review consisted of a sequence of word tokens &
    $\mathbb{R}^{n \times d_w}$ \\
    $n$ & Number of word tokens in $R_i$ & $\mathbb{R}^{100}$ \\
    $t_i^{(j)}$ & $j$-th word in $R_i$ & $\mathbb{R}^{d_w}$ \\
    $A$ & Predefined aspect categories & $\mathbb{R}^{N}$ \\
    $A_{R_i}$ & The set of aspects present in $R_i$ & $\mathbb{R}^{K}(K \leq
    N)$\\
    $A_{R_i}^{(k)}$ & $k$-th aspect in $A_{R_i}$ & $\mathbb{R}^{d_w}$ \\
    $y_{A_{R_i}^{(k)}}$ & Sentiment polarity of $A_{R_i}^{(k)}$ &
    $\mathbb{R}^{3}$ \\
    $y_{R_i}$ & Star rating of $R_i$ & $\mathbb{R}^{3}$ \\
    $M$ & Model & - \\
    $\hat{A}_{R_i}$ & Prediction of $A_{R_i}$ & - \\
    $\hat{y}_{A_{R_i}^{(k)}}$ & Prediction of $y_{A_{R_i}^{(k)}}$ &
    $\mathbb{R}^{3}$ \\
    $\hat{y}_{R_i}$ & Prediction of $y_{R_i}$ & $\mathbb{R}^{3}$ \\
    $\mathbf{X}_i$ & Input sequence & $\mathbb{R}^{n \times d_w}$ \\
    $\mathbf{z}_i$ & Sentence embedding of $\mathbf{X}_i$ (pooler\_output of
    BERT) & $\mathbb{R}^{d_w}$ \\
    $\mathbf{T}$ & Aspect embedding matrix & $\mathbb{R}^{N \times d_w}$ \\
    $\mathbf{T}_k$ & Embedding of $k$-th aspect & $\mathbb{R}^{d_w}$ \\
    $\mathbf{r}_i$ & Reconstructed sentence embedding & $\mathbb{R}^{d_w}$ \\
    $\mathbf{p}_i$ & Weight vectors of $K$ aspect embeddings (aspect importance)
    & $\mathbb{R}^{N}$ \\
    $L(\theta_{ACD})$ & Loss function of ACD task (Module 1) & - \\
    $\lambda_{ACD}$ & Weight of regularization term & - \\
    $U(\theta)$ & Regularization term & - \\
    $\mathbf{n}_j$ & Sentence embedding of negative sample & $\mathbb{R}^{d_w}$ \\
    $\mathbf{h}_i^{(j)}$ & hidden state of $j$-th word (last\_hidden\_state of
    BERT) & $\mathbb{R}^{d_w}$ \\
    $\mathbf{w}_i^{(j)}$ & Sentiment prediction vector of $j$-th word &
    $\mathbb{R}^{d_w \times 3}$ \\
    $d_k^{(j)}$ & Distance between $j$-th word and $k$-th aspect & - \\
    $a_k^{(j)}$ & Attention weight of $j$-th word towards $k$-th aspect  & - \\
    $L(\theta_{RP})$ & Loss function of RP task (Module 2) & - \\
    $\lambda$ & Weight of $L(\theta_{ACD})$ & - \\
    $L(\theta)$ & Overall loss function & - \\

\bottomrule
\end{tabular}
} \caption{\label{tab:variables} Description of variables in our formulation.}
\end{table}

\subsection{Constituent Components}

\paragraph{Aspect-Category Detection}

In the ACD task, there are $N$ predefined aspect categories (hereafter
``aspects''): $A = \{A_1, A_2, \dots, A_N\}$. The set of aspects present in
$R_i$ is defined as: $A_{R_i} = \{A_{R_i}^{(1)}, A_{R_i}^{(2)}, \dots ,
A_{R_i}^{(K)}\}$,  where $K \leq N$. To train unsupervised ACD models, the
required training data is simply $\mathbf{R}$.

\paragraph{Aspect-Category Sentiment Analysis} 

For a given review $R_i$ and one of its aspects $A_{R_i}^{(k)}$, the goal of
ACSA is to predict the polarity of the aspect: $\hat{y}_{A_{R_i}^{(k)}}$. Aspect
polarity is typically binary (\textit{positive} or \textit{negative}) or
categorical (with a third option of \textit{neutral}). Supervised ACSA models
require review-aspect-polarity triples: $\{R_i, (A_{R_i}^{(k)},
y_{A_{R_i}^{(k)}})_{k=1}^K\}_{i=1}^{\vert \mathbf{R} \vert}$. In the case of
multi-aspect ACSA, multiple aspects are present in each review, and therefore,
ACSA requires $K \times \vert \mathbf{R} \vert$ labels for each review, a factor
of $K$ larger than in RP.

\paragraph{Rating Prediction}
Given a review $R_i$, RP aims to predict the star rating $\hat{y}_{R_i}$.
Supervised RP models requires review-sentiment tuples: $\{(R_i,
y_{R_i})\}_{i=1}^{\vert \mathbf{R} \vert}$.

\subsection{Unified Sentiment Analysis Training}
\label{ssec:dle2esa}

Typically, ACD, ACSA, and RP are considered standalone tasks. With unified sentiment analysis, we
propose a unified approach, where with training data of only RP labels, a model
can output present aspects (ACD), the sentiment of those aspects (ACSA), and an
overall document-level sentiment score (RP). This approach uses training labels
from a single task (RP) to efficiently learn multiple distinct sentiment
analysis tasks. More specifically, for a model $M$, the training data required
is the same as the RP task: $\{(R_i, y_{R_i})\}_{i=1}^{\vert \mathbf{R} \vert}$.
At run-time, the model provides three outputs for a new review $R_i$: (1) The
predicted aspects present in the review ($\hat{A}_{R_i}$), (2) the sentiment
polarity of each identified aspect ($\hat{y}_{A_{R_i}^{(k)}}  \forall
A_{R_i^{(k)}} \in \hat{A}_{R_i}$), and (3) the overall sentiment prediction for
the review ($\hat{y}_{R_i}$).

\section{Distantly Supervised Pyramid Network}
\begin{figure}[h!]
\centering
\includegraphics[width=0.75\columnwidth]{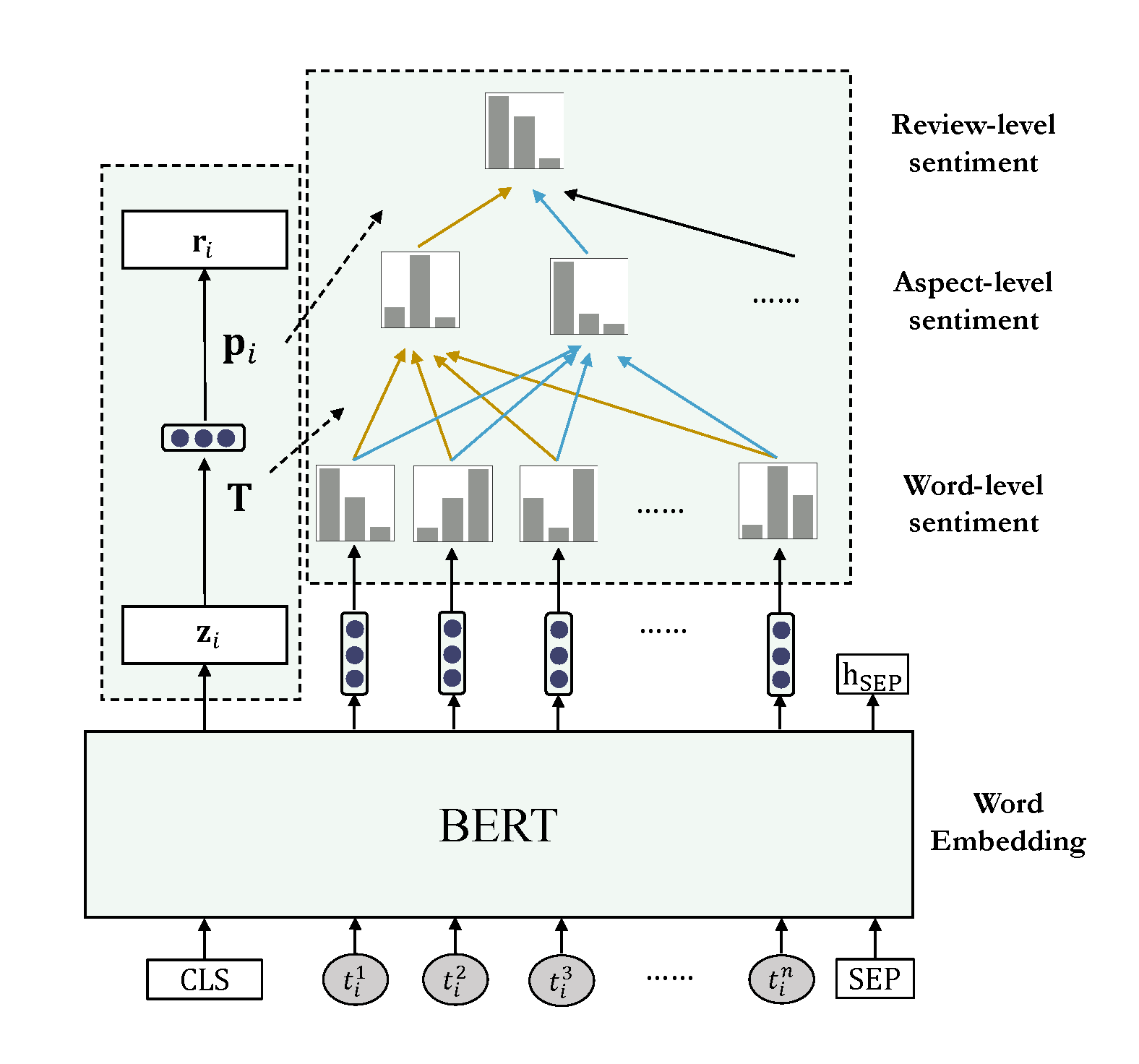}
\caption{Overall architecture of DSPN. Aspect embedding matrix $\mathbf{T}$ is
used to calculate the distance between words and aspects, which is regarded as
the word-level attention weights for each aspect. Aspect importance
$\mathbf{p}_i$ is learned by Module 1 and is used as the attention weights of
aspects.}
\label{FigArchitecture}
\end{figure}

In this section, we describe DSPN, our novel implementation for unified sentiment analysis. Figure \ref{FigArchitecture} illustrates the DSPN model architecture.

\subsection{Module 1: Aspect-Category Detection}
\label{ssec:module1}

As a foundation for our ACD module, we draw on an existing autoencoder-style
network~\cite[ABAE, ][]{he2017unsupervised}, but enhance it to 
address two limitations of the model.
First, ABAE uses Word2Vec for word embeddings, which constrains its performance due to the static embeddings and limits the model's ability to capture contextual nuances. 
In DSPN, we replaced Word2Vec with BERT to better capture semantic relationships with context-aware embeddings. 
Second, ABAE requires manually mapping learned aspect embeddings to specific aspects via similar words in the embedding space and estimating the number of aspects via clustering.  
Therefore, we redesigned the aspect embedding matrix to match the number of aspects in the dataset, eliminating the need for manual aspect extraction. 
In DSPN, the model learns aspect-specific weights, which, combined with a thresholding mechanism, allows for more accurate and automated aspect identification.\footnote{See Appendix \ref{sec:dspnw2v} for an empirical analysis of our design decisions and their impact.} 
We next elaborate on the details of our novel ACD module.

For a review $R_i$, the input sequence
$\mathbf{X}_i$ is constructed as $\{[CLS], t_i^{(1)}, t_i^{(2)}, \dots,
t_i^{(n)}, [SEP]\}$. We use BERT~\citep{devlin2019bert} to generate embeddings
$\mathbf{z}_i$ for each sample.
To generate aspect embeddings, we first set the aspect and keyword map
dictionary for each aspect.\footnote{There are $N$ predefined aspects in ACD
task, and many prior works have identified the representative words for each one
of them~\citep{bu2021asap, wang2010latent}. For example, ``staff,'' ``customer,''
and ``friendly'' can be the representative words for ``Service'' aspect. Based
on this, we proposed first constructing a sentence that contains top
representative words, then using the embedding of this sentence as the initial
embedding for the aspect.} Then, for each aspect, we use BERT to encode the
sentence composed of key words related to the aspect and obtain its output as
the initial embedding of the aspect. In this way, we initialize the aspect
embedding matrix $\mathbf{T}$. Lastly, Module 1 performs sentence reconstruction
at the aspect level through a linear layer:

\begin{align}
    \mathbf{z}_i = \text{BERT}&(\mathbf{X}_i) \label{eq3}\\
    \mathbf{p}_i = \text{softmax}(&\mathbf{W_1}\cdot \mathbf{z}_i + \mathbf{b_1}) \label{eq4} \\
    \mathbf{r}_i = \mathbf{T}^{\top}\cdot & \mathbf{p}_i \label{eq5} 
\end{align}

where $\mathbf{r}_i$ is the reconstructed sentence embedding and $\mathbf{p}_i$
is the aspect importance vector.

The loss function for the unsupervised Module 1 is defined as a hinge loss to
maximize the inner product between the input sentence embedding and its
reconstruction while minimizing the inner product between the input sentence
embedding and randomly sampled negative examples:

\begin{align}
L(\theta_{\text{ACD}}) &= \sum_{R_i \in \mathbf{R}} \sum_{j=1}^m \phi_{R_i,j} + \lambda_{\text{ACD}} U(\theta)
    \label{eq6} \\
    \phi_{R_i,j} &= \max(0, 1-\mathbf{r}_i\mathbf{z}_i + \mathbf{r}_i\mathbf{n}_j) \label{eq7}
\end{align}

where $\mathbf{n}_j$ represents BERT embeddings for the negative sample, $\mathbf{n}_j = \text{BERT}(\mathbf{X}_j)$, and $U(\theta)$ represents
the regularization term to encourage unique aspect
embeddings~\citep{he2017unsupervised}. The aspect embedding matrix $\mathbf{T}$
and aspect importance vector $\mathbf{p}_i$ are inputs for attention calculation
in DSPN's pyramid network (Module 2).

\subsection{Module 2: Pyramid Sentiment Analysis}

Module 2 is based on the intuition that the sentiment of a review is an
aggregation of the sentiments of the aspects contained in the
review~\citep{bu2021asap}. In addition, the sentiment of an aspect is an
aggregation of the sentiments of the words indicating that aspect, forming a
three-layer structure. We propose using a pyramid network to capture this
structure, and we can use easy-to-obtain RP ratings as training labels. In the
word sentiment prediction layer, we use the hidden vector of each word output by
BERT to obtain word representations, where $\mathbf{h}_i^{(j)}$ is the
representation of the $j$-th word. We use two fully connected layers to produce
a word-level sentiment prediction vector:

\begin{equation} 
\mathbf{w}_i^{(j)} = \mathbf{W_3} \cdot \text{ReLU}(\mathbf{W_2} \cdot \mathbf{h}_i^{(j)}+\mathbf{b_2}) + \mathbf{b_3} \label{eq8}
\end{equation}


Then, we can calculate the similarity of words and aspects using the word
representations and the aspect embedding matrix $\mathbf{T}$ output by Module 1.
This similarity will be treated as the attention weights of words for the
aspect. When predicting aspect-level sentiment, for the $k$-th aspect, the
sentiment $\hat{y}_{A_{R_i}^{(k)}}$ is computed as:

\begin{align}
    d_{k}^{(j)} &= \mathbf{T}_k^{\top} \cdot \mathbf{h}_i^{(j)} \label{eq9} \\
    a_k^{(j)} =& \frac{\exp(d_k^{(j)})}{\sum_{m=1}^n \exp(d_k^{(m)})} \label{eq10} \\
    \hat{y}_{A_{R_i}^{(k)}} = &\text{sof}\text{tmax}(\sum_{j=1}^n \mathbf{w}_i^{(j)} a_k^{(j)}) \label{eq11} 
\end{align}


Finally, the review-level sentiment $\hat{y}_{R_i}$ is computed by: 

\begin{equation}
\hat{y}_{R_i} = \text{softmax}(\hat{y}_{A_{R_i}} \cdot \mathbf{p}_i) \label{eq12}
\end{equation}

Here, $\mathbf{p}_i$ is the aspect importance vector output by Module 1 (\S
\ref{ssec:module1}), which is regarded as the attention weights of aspects in a
review. $\hat{y}_{A_{R_i}}$ is the matrix concatenation of aspect-level
sentiments in the review.

\subsection{Loss}

For the RP task, as each prediction is a 3-class classification problem, the
loss function is defined by the categorical cross-entropy between the true label
and the model output: 

\begin{equation}
L(\theta_{\text{RP}}) = -\sum_{i} y_{R_i} \cdot \log(\hat{y}_{R_i}) \label{eq13}
\end{equation} 

We jointly train DSPN for RP and ACD by minimizing the combined loss function: 

\begin{equation}
L(\theta) = \lambda L(\theta_{\text{ACD}}) + L(\theta_{\text{RP}}) \label{eq14}
\end{equation} 

where $\lambda$ is the weight of ACD loss. Although no direct supervision is
required for ACSA,  due to the construction of DSPN, the model inherently learns
aspect sentiment predictions.

\subsection{Inference}

For any new text input (e.g., restaurant or hotel review)
$R_i$, DSPN generates three outputs: a distribution over aspects
($\mathbf{p}_i$), estimated sentiment for each aspect
($\hat{y}_{A_{R_i}^{(k)}}$), and an overall sentiment prediction
($\hat{y}_{R_i}$).

\section{Experiments}

\subsection{Datasets}
As described above, DSPN can model ACD, ACSA, and RP with only RP labels.
However, in order to benchmark DSPN
across the three unified sentiment analysis tasks, we need datasets with ACD, ACSA, and RP labels.
Such datasets are not common in the literature. After a thorough review of
available data, we only found two datasets sufficient for our benchmarking.
To validate DSPN as an efficient and effective model for unified sentiment
analysis, we experiment with these two datasets: one in English (TripDMS) and one in
Chinese (ASAP). Dataset details are provided in Table
\ref{tab:DataDetails}.

TripDMS is an English-language hotel review dataset from
Tripadvisor.com~\citep{wang2010latent, yin2017document}, with RP labels on a
5-star scale and ACSA labels (\textit{positive}, \textit{negative},
\textit{neutral}) for seven aspects: {Value, Room, Location, Cleanliness,
Check-in, Service, Business}. ASAP is a Chinese-language restaurant review
dataset from a leading e-commerce platform in China~\citep{bu2021asap}. It
includes RP labels on a 5-star scale and ACSA labels (\textit{positive},
\textit{negative}, \textit{neutral}) for aspects identified in the review
text~\citep{pontiki2016semeval}. For ACSA, sentiment is aggregated at the entity
level for five aspects: {Food, Price, Location, Service, Ambience}, determined
by majority vote. We reiterate that, to the best of our knowledge, these are the only two publicly available datasets with both RP and ACSA labels for evaluating performance. Hence, we are unable to validate our proposed model on other datasets at this time.

\label{sec:datasets}

\begin{table}[h!]
\centering
\footnotesize
\scalebox{0.9}{
\begin{tabular}{ccccccccccccc}
\toprule
\multirow{2}{*}{\textbf{Dataset}} & \multirow{2}{*}{\textbf{Language}} &
\multirow{2}{*}{\textbf{MA}} & \multirow{2}{*}{\textbf{MAS}} &
\multirow{2}{*}{\textbf{Split}} & \multirow{2}{*}{\textbf{Reviews}} &
\multicolumn{3}{c}{\textbf{Overall Sentiment}} &
\multicolumn{4}{c}{\textbf{Aspect Sentiments}} \\
\cmidrule(lr){7-9} \cmidrule(lr){10-13} ~ & ~ & ~ & ~ & ~ & ~ & \textbf{Pos.} &
\textbf{Neu.} & \textbf{Neg.} & \textbf{Pos.} & \textbf{Neu.} & \textbf{Neg.} &
\textbf{NaN} \\
\midrule
    \multirow{3}{*}{TripDMS} & \multirow{3}{*}{English} & \multirow{3}{*}{100\%}
    & \multirow{3}{*}{100\%} & Train & 23,515 & 8,998 & 5,055 & 9,462 & 64,984 &
    34,200 & 43,391 & 22,030 \\
    ~ & ~ & ~  & ~& Val & 2,939 & 1,161 & 613 & 1,165 & 8,174 & 4,245 & 5,349 &
    2,805 \\
    ~ & ~ & ~ & ~ & Test & 2,939 & 1,079 & 647 & 1,213 & 8,002 & 4,355 & 5,437 &
    2,779 \\ 
    \midrule
    \multirow{3}{*}{ASAP} & \multirow{3}{*}{Chinese} & \multirow{3}{*}{95.97\%}
    & \multirow{3}{*}{63.85\%} & Train & 36,850 & 29,132 & 5,241 & 2,477 &
    77,507 & 27,329 & 17,299 & 62,115 \\
    ~ & ~ & ~ & ~ & Val & 4,940 & 3,839 & 784 & 317 & 10,367 & 3,772 & 2,373 &
    8,188 \\
    ~ & ~ & ~ & ~ & Test & 4,940 & 3,885 & 717 & 338 & 10,144 & 3,729 & 2,403 &
    8,424 \\
\bottomrule
\end{tabular}
} \caption{\label{tab:DataDetails} Statistics of the datasets. \textbf{MA} is
the percentage of multi-aspect instances in the dataset and \textbf{MAS} is the
percentage of multi-aspect multi-sentiment instances.}
\end{table}

\subsection{Benchmarking and Hyperparameter Settings}
\label{ssec:hyperparams}

Recall that DSPN aims to achieve higher efficiency and accuracy in performing
unified sentiment analysis through distant supervision. Therefore, to test its
performance, we compare DSPN's performance to that of existing ACD, ACSA, and RP
models.
Specifically, for ACD, we compare DSPN with the unsupervised
ABAE~\citep{he2017unsupervised}. To ensure a fair comparison with DSPN, we
replaced ABAE's encoder with BERT and updated the aspect embedding matrix
$\mathbf{T}$ initialization. We call this ABAE-BERT and report its performance.
Following previous work~\citep{ruder2016insight, ghadery2019licd}, we use
thresholding to assign aspects based on probability, selecting the threshold
that yields the best performance ($1e^{-4}$). ACD is evaluated using F1 score.
For ACSA, we benchmark DSPN against several strong supervised models,
incorporating non-BERT models (GCAE~\citep{xue2018aspect} and
End2end-LSTM/CNN~\citep{schmitt2018joint}) 
as well as BERT-based models 
(ACSA-Generation~\citep{liu2021solving}). ACSA is evaluated using accuracy.
For RP, a text classification task, we compare DSPN with BERT fine-tuning
strategies~\citep{sun2019fine}: BERT-Feat, BERT-FiT, and BERT-ITPT-FiT. We
convert the 5-star RP ratings into three classes (Negative, Neutral, Positive)
and evaluate using accuracy.

We performed all of our experiments on Nvidia H100 GPUs.
We set the word embedding size to 768~\citep{devlin2019bert} and the
maximum sentence length to 100~\citep{li2020multi}.
For other parameters, we conduct a grid search across a range of 
values and evaluate
on a held-out dataset to identify those that best fit the task.
Specifically, we run grid search on the following parameter-value combinations: 
batch size (16, 32), training epochs (5, 10), and learning rate (2e-5, 3e-5, 5e-5).


\section{Results}
\subsection{Overall Performance}
\label{ssec:overall_analyses}

To compare DSPN with existing models, we evaluate two pipeline approaches: a
\textit{high-performance} pipeline using the best-performing model for each task
and a \textit{high-efficiency} pipeline using the most efficient model in terms
of parameters. Our benchmarking results are shown in Tables
\ref{tab:highperformance} and \ref{tab:highefficiency}. The efficiency metric is
the sum of the separate models' parameters; the performance metric
reflects the individual performance of each model on the three
tasks.\footnote{Results for all benchmarking models are presented in Appendix
\ref{sec:ab} for completeness.}


\begin{table}[t]
	\centering
    \scriptsize
    \scalebox{0.95}{
	\begin{tabular}{clcccccc}
		\toprule 
		& & & \bf Efficiency & &&\bf Performance &  \\
		\cmidrule(lr){3-5} \cmidrule(lr){6-8} &                & \bf Parameters
& \bf Labels &\bf Training Time & \bf ACD  & \bf ACSA  & \bf RP   \\
&& (Millions) & (Thousands)  & (Minutes) & (F1) & (Acc) & (Acc) \\
\midrule 
		\multirow{6}{*}{TripDMS} & ABAE-BERT (ACD) & 109.49  & 0.00 
		& 35.21 (0)  & 92.72 (0)   &      &
		\\
		& ACSA-Generation (ACSA) & 255.28     & 142.58 & 59.51 (0.15)
		&                 & 64.21 (0.18)             &                 \\
		& BERT-ITPT-FiT (RP)   & 112.71    & 23.52  & 35.34 (0.14)
		&        &      & 72.94 (0.42)            \\
		& Pipeline      & 477.48          & 166.10 & 130.06 (0.34)
		& 92.72 (0)   & \textbf{64.21 (0.18)}   & \textbf{72.94 (0.42)}   \\
		& DSPN (Ours)    & \textbf{109.50}      & \textbf{23.52}
		& \textbf{38.66 (0.01)}  & 92.72 (0)     & 50.00 (0.71)
		& 71.70 (0.41)            \\
		\cmidrule(lr){3-5} \cmidrule(lr){6-8} & Delta    & -77.07
		& -85.84 & -70.28     & 0.00            & -22.13     &
		-1.70            \\
		\midrule 
		\multirow{6}{*}{ASAP}   & ABAE-BERT (ACD)  & 102.28     & 0.00
		& 54.48 (0.07)    & 79.44 (0)   &    &
		\\
		& ACSA-Generation (ACSA)       & 225.28  & 122.14 & 51.28 (1.25)
		&                 & 75.92 (0.17)    &        \\
		& BERT-ITPT-FiT (RP)         &  103.50       & 36.85  & 38.44 
 (0.09)
		&                 &                  & 81.58(1.30)          \\
		& Pipeline             & 431.06         & 158.99 & 144.20 (0.57)
		& 79.44 (0)           & \textbf{75.92 (0.17)}           & \textbf{81.58 (1.30)}
		\\
		& DSPN (Ours)                & \textbf{102.28}             & \textbf{36.85}
		& \textbf{60.51 (0.04)}                      & 79.44 (0)           & 64.63 (0.02)
		& 80.85 (0.11)            \\
		\cmidrule(lr){3-5} \cmidrule(lr){6-8} & Delta                & -76.27
		& -76.82 & -58.04     & 0.00           & -14.87         & -0.89 \\
		\bottomrule          
	\end{tabular}
    }
	\caption{Comparison between DSPN and a high-performance pipeline approach to
	unified sentiment analysis. For efficiency comparisons, negative deltas
	indicate improvements over the pipeline; for performance, positive deltas
	indicate improvements over the pipeline. ACD, ACSA, and RP refers to aspect-category detection, aspect-category sentiment analysis, and rating prediction respectively. For training time and performance metrics we report the mean and variance over five runs with different random seeds.}
\label{tab:highperformance}
\end{table}

\begin{table}[t]
	\centering
    \scriptsize
    \scalebox{0.95}{
	\begin{tabular}{clcccccc}
		\toprule 
& & & \bf Efficiency & &&\bf Performance &  \\
\cmidrule(lr){3-5} \cmidrule(lr){6-8} &                & \bf Parameters & \bf
Labels &\bf Training Time & \bf ACD  & \bf ACSA  & \bf RP   \\
&& (Millions) & (Thousands)  & (Minutes) & (F1) & (Acc) & (Acc) \\
\midrule 
		\multirow{6}{*}{TripDMS} & ABAE (ACD)      & 14.81             & 0.00     &
		36.20 (0.08)                    & 91.22 (0)           &                  &
		\\
		& GCAE (ACSA)     & 15.27            & 142.58 & 7.64 (0.01)                     &
		& 57.97 (1.43)         &                 \\
		& BERT-Feat (RP) & 109.49       & 23.52  & 3.48 (0)                   &
		&                  & 70.04 (0.43)            \\
		& Pipeline  & 139.57           & 166.10 & 47.32 (0.07)
		& 91.22 (0)          & \textbf{57.97 (1.43)}            & 70.04 (0.43)         \\
		& DSPN (Ours)     & \textbf{109.50}          & \textbf{23.52}  & \textbf{38.66 (0.01)}
		& \textbf{92.72 (0)}            & 50.00 (0.71)         & \textbf{71.70 (0.41)}
		\\
		\cmidrule(lr){3-5} \cmidrule(lr){6-8} & Delta     & -21.54           &
		-85.84 & -18.30                  & 1.64            & -13.75            & 2.37
		\\
		\midrule 
		\multirow{6}{*}{ASAP}   & ABAE (ACD)     & 14.81       & 0.00     & 37.42 (0.49)
		& 79.41 (0)          &                  &                 \\
		& GCAE (ACSA)     & 15.27   & 122.14 &  6.13 (0.05)     &
		& 74.50 (0.08)           &                 \\
		& BERT-Feat (RP) & 102.27        & 36.85 & 2.06 (0)                  &
		&                  & 78.61 (0)        \\
		& Pipeline  & 132.35          & 158.99 & \textbf{45.61 (0.69)}
		& 79.41 (0)          & \textbf{74.50 (0.08)}       & 78.61 (0)
		\\
		& DSPN (Ours)     & \textbf{102.28}    & \textbf{36.85}  & 60.51 (0.04)
		& \textbf{79.44} (0)          & 64.63 (0.02)          & \textbf{80.85 (0.11)}
		\\
		\cmidrule(lr){3-5} \cmidrule(lr){6-8} & Delta     & -22.72           &
		-76.82 & 32.67     & 0.04            & -13.25            & 2.85
		\\
		\bottomrule       
	\end{tabular}
    }
	\caption{Comparison between DSPN and a high-efficiency pipeline approach to
	unified sentiment analysis. For efficiency comparisons, negative deltas
	indicate improvements over the pipeline; for performance, positive deltas
	indicate improvements over the pipeline. ACD, ACSA, and RP refers to aspect-category detection, aspect-category sentiment analysis, and rating prediction respectively. For training time and performance metrics we report the mean and variance over five runs with different random seeds.}
\label{tab:highefficiency}
\end{table}

We conducted a Wilcoxon signed rank test to compare the scores over five runs with different seeds to determine whether one sequence of scores was significantly higher than the other. 
High performance pipeline performance was significantly higher than DSPN for ACSA on both TripDMS and ASAP datasets ($p < 0.05$).
For TripDMS, pipeline RP performance was higher ($p < 0.1$).
Scores were not significantly different for ACD on either dataset.
This suggests that DSPN's performance is comparable with our high-performance pipeline on ACD and RP.
Comparing DSPN to the high-efficiency pipeline, DSPN performance is significantly higher for ACD and RP for both datasets ($p < 0.05$ in all cases except ACD ASAP: $p < 0.1$), while the pipeline performance is higher ($p < 0.05$) for ACSA (both datasets).
The ACSA results are expected because of the supervised nature of the pipeline. 
However it is encouraging to see DSPN performance improvements over the pipeline for ACD and RP. 

DSPN's performance on ACSA is lower than the supervised benchmarks. 
This is to be expected as DSPN's only supervision is RP labels. From an
efficiency point of view, ACSA models require 142,575 labels on TripDMS to learn
one task (ACSA), while DSPN only requires 23,515 labels (84\% fewer) to learn
three tasks. Therefore, by using DSPN, we see a tradeoff of an 84\% reduction in
labels for a prediction performance that is only 22\% lower than the
best-performing supervised model for ACSA. Similarly, for ASAP, we observe a
label-performance tradeoff where for a 70\% reduction in labels, DSPN
performance is only 15\% lower than the best-performing supervised model for ACSA. 
What's more, DSPN outperforms the fully-supervised End2end-CNN baseline model.

Our single-task benchmarks set the ``upper bound'' of performance for the task
when given a fully labeled dataset. They can be considered target values for
future unified sentiment analysis models. However, if only RP labels exist for a given dataset, then
DSPN is the only method for learning all three tasks. 
Considering that DSPN does not use any aspect-level labels, the fact that the
effectiveness of DSPN is comparable to supervised models on the ACSA task is
empirical validation of the unified sentiment analysis in general and the DSPN architecture in
particular.

\begin{figure}[h!]
    \small
	\centering
	\includegraphics[width=0.95\textwidth]{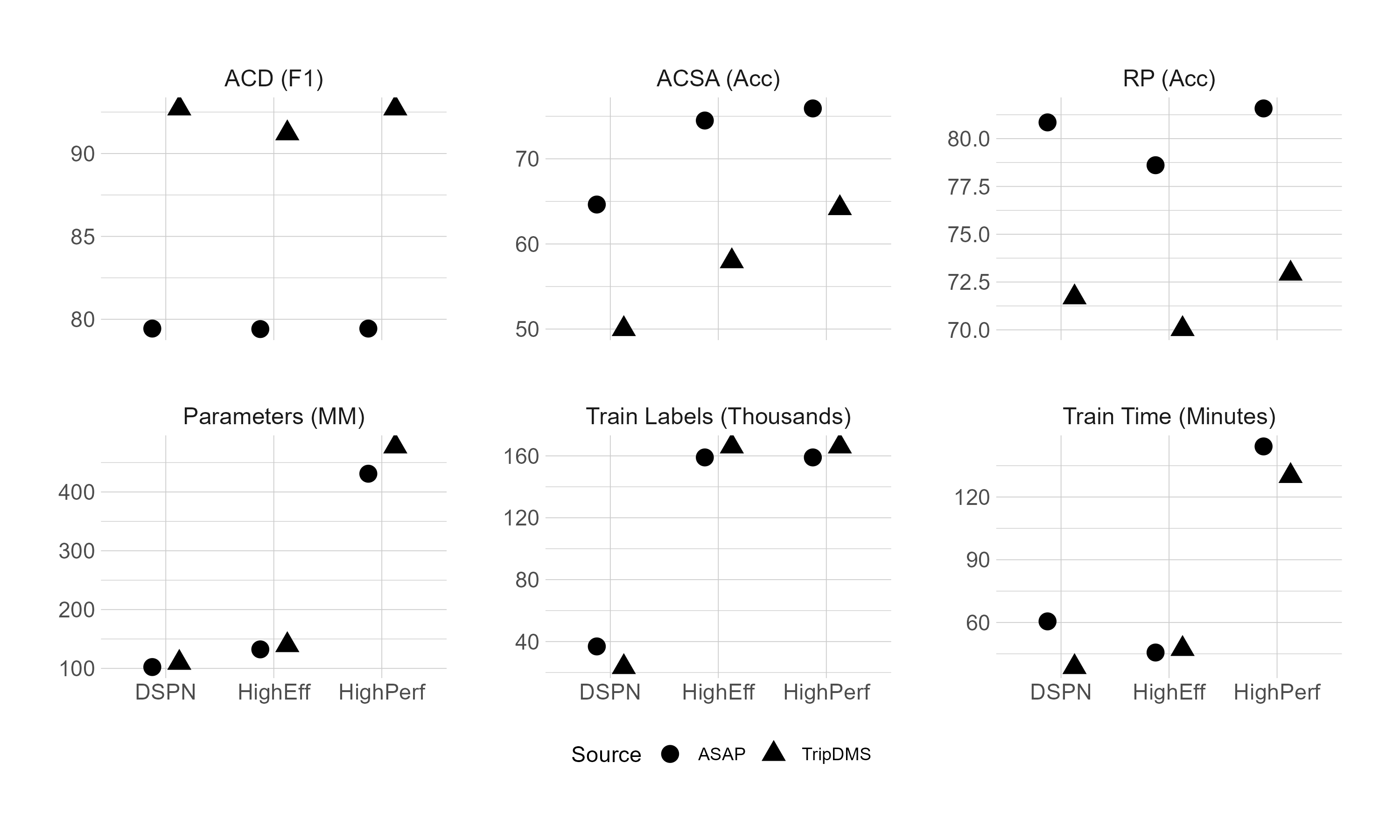}
	\caption{Comparing DSPN to the two pipelines. In the top row, higher values are better. In the bottom row, lower values are better.}
	\label{fig:pipelinesPlotComb}
\end{figure}

To better visualize the comparison between DSPN and the two pipelines, we
plotted each relevant metric in Figure \ref{fig:pipelinesPlotComb}. From a
managerial perspective, models are often incorporated into broader
pipelines~\citep{lalor2024should}. Therefore, considering a holistic view and
analyzing the metrics together is key for decision makers. 

\subsection{Quality Analysis}

\subsubsection{Case Study}

\begin{figure*}[h!]
\centering
\begin{subfigure}{1\textwidth}
\centering
\includegraphics[width=0.9\textwidth]{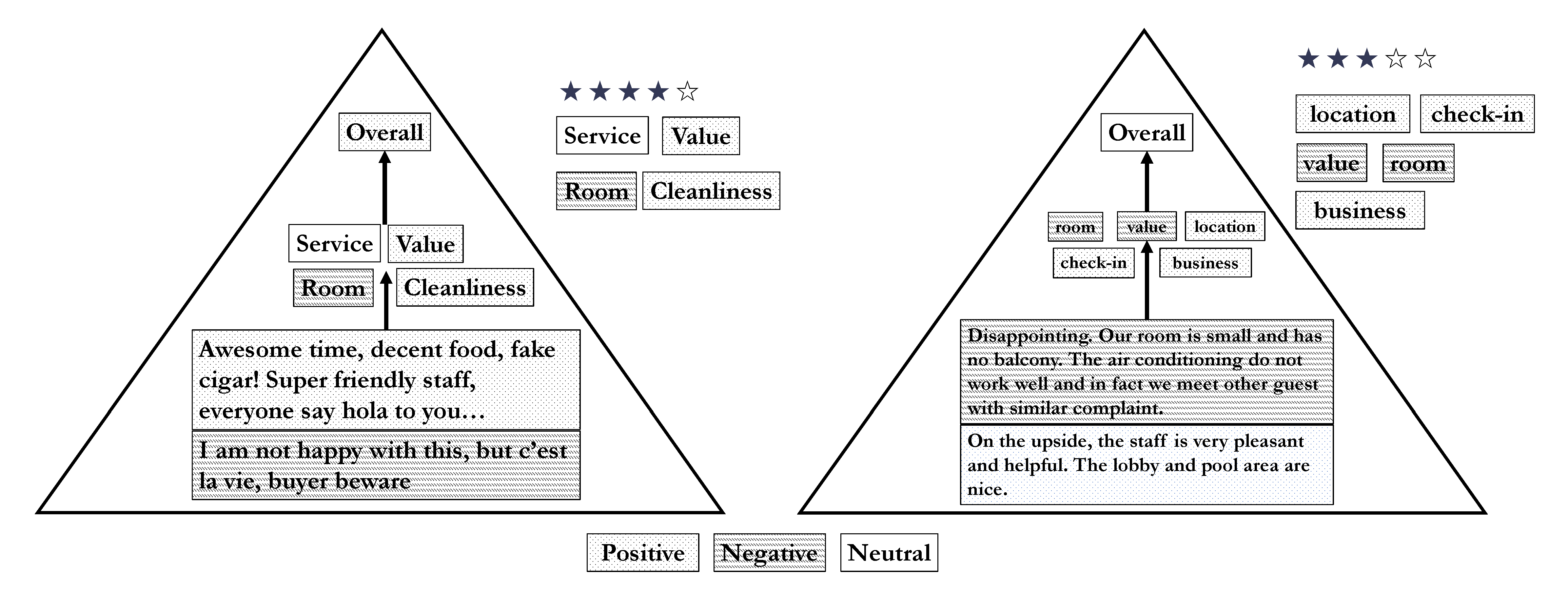}
\caption{\label{correct_case}}
\end{subfigure}
\begin{subfigure}{1\textwidth}
\centering
\includegraphics[width=0.95\textwidth]{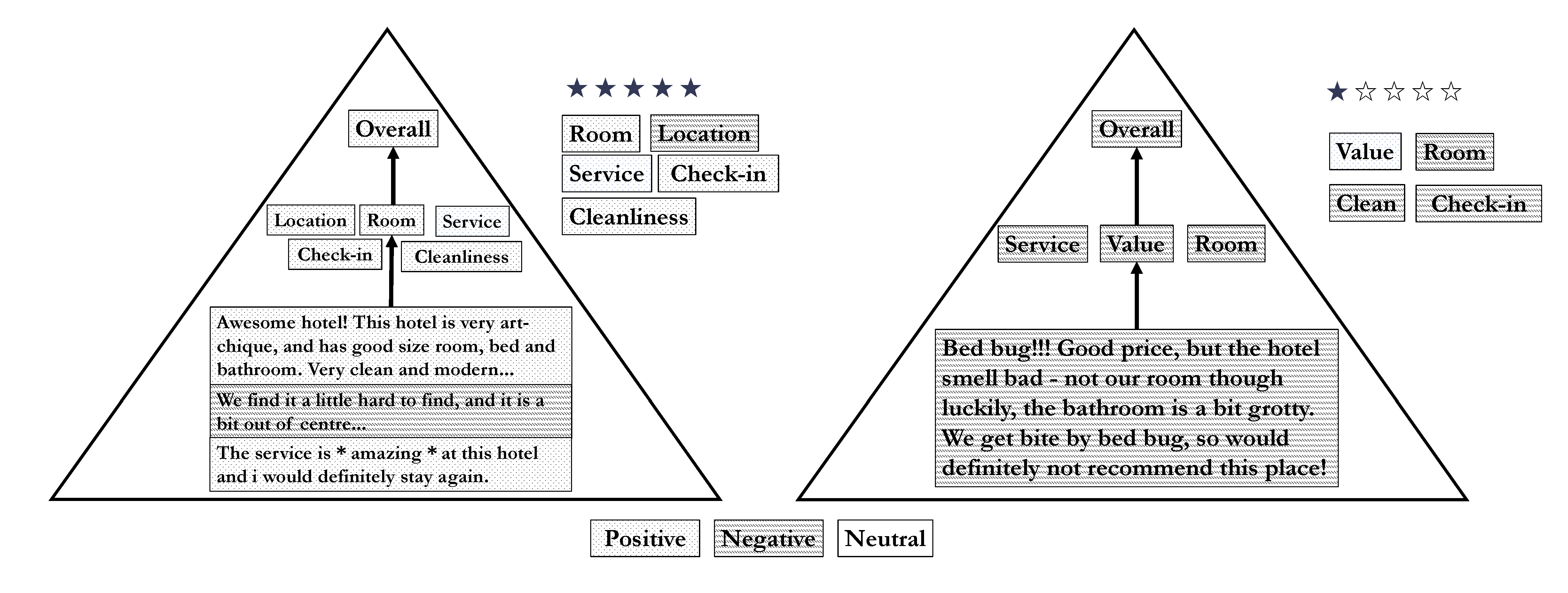}
\caption{\label{error_case}}
\end{subfigure}
\caption{Case studies of correct predictions (\ref{correct_case}) and incorrect
predictions (\ref{error_case}). True RP and ACSA labels are outside of the
pyramid; DSPN's predictions are within the pyramid. For space, we show a portion
of the review.}
\label{cases}
\end{figure*}

In order to visualize and analyze DSPN's performance, we first take two reviews
from TripDMS as examples~(Figure \ref{correct_case}). For each example, the
trained DSPN model takes the review text as input and first outputs word-level
sentiment predictions. Then, DSPN (i) identifies aspect keywords via a word
attention calculation, (ii) obtains the aspect importance, (iii) calculates
aspect-level sentiment through the sentiments of the key words, and lastly (iv)
combines aspect sentiment with aspect importance to predict the final
review-level sentiment (``Overall'' in Figure \ref{cases}).

For case 1 in Figure \ref{correct_case}, DSPN correctly labels the review as
positive and also correctly identifies and labels the \textit{Service},
\textit{Value}, \textit{Room}, and \textit{Cleanliness} aspects with no
aspect-level annotations. For case 2, DSPN gives correct predictions on word-,
aspect-, and review-level sentiments.

\subsubsection{Error Analysis}

To exemplify errors in DSPN, we then examine two examples of error cases from
TripDMS in Figure \ref{error_case}. We find that extreme star rating labels
sometimes influence DSPN. For example, for case 1 in Figure \ref{error_case},
DSPN gives correct word-level sentiments but tends to give positive predictions
at the aspect level due to the overall 5-star rating. Similarly, for case 2,
DSPN gives negative predictions on all three levels due to a 1-star rating. This
is to be expected as DSPN's only supervision is star rating labels. Addressing
this disconnect is a promising area for future unified sentiment analysis model development.

\section{Additional Analyses}
\label{sec:analyses}

\subsection{Budget Constraint Experiment}

For a more direct comparison between DSPN and the supervised ACSA models, we
designed a budget-constraining experiment. 
Specifically, we randomly selected ACSA labels for TripDMS and ASAP so that the
supervised models have the same training label size as DSPN.
In this setting, DSPN's performance is closer to the supervised models'
performance~(Table \ref{tab:Budget}). 
In particular, DSPN outperforms End2end-CNN on both dataset. 
Overall, the supervised models still outperform DSPN, but this is to be expected given that the labels used for training are ACSA labels. Recall that DSPN still performs RP and ACD on top of these ACSA results, while the benchmark models only learn ACSA. DSPN is trained
to perform RP but is also able to perform ACSA in a way that is comparable to
these supervised models under the same budget constraint.

\begin{table}[h!]
	\centering
	\begin{tabular}{lcc}
		\toprule
		\textbf{Model} & \textbf{TripDMS} & \textbf{ASAP} \\
		\midrule
		End2end-LSTM & 52.7 & 57.5 \\
		End2end-CNN & 41.4 & 43.2 \\
		GCAE & 54.0 & 70.1 \\
		ACSA-Generation & 60.2 & 72.3 \\
		DSPN (Ours) & 50.0 & 64.6 \\    
		\bottomrule
	\end{tabular}
	\caption{\label{tab:Budget} ACSA accuracy results when all models are trained with
	the same amount of data.}
\end{table}

\subsection{Comparing Weakly Supervised ACSA Models}
\label{ssec:weaklysupervised}

We have shown DSPN's effectiveness using two datasets that include both
review-level star rating labels (for RP) and aspect-level sentiment annotations
(for ACSA). 
What's more, there are methods in the literature that take a weakly supervised approach
to ACSA, such as JASen~\citep{huang2020weakly} and AX-MABSA~\citep{kamila2022ax}.
However, current semi-supervised ACSA models have certain limitations. For example, JASen and AX-MABSA are unable to model neutral sentiment polarity.

Therefore, to ensure fair and meaningful comparisons, we selected compatible datasets and conducted comprehensive benchmarking against these methods.
Specifically, we compare DSPN to semi-supervised models on four ACSA datasets:
Rest-14~\citep{pontiki-etal-2014-semeval},
Rest-15~\citep{pontiki2015semeval}, Rest-16~\citep{pontiki2016semeval} and
MAMS~\citep{jiang2019challenge}.
To enable DSPN to run on such datasets without star rating labels, we use the aggregate value of aspect
ratings as the training labels instead of the star rating labels given by users.
Table \ref{tab:noRP} shows that DSPN performs comparably to JASen. This result
indicates that RP is not simply an average over ACSA labels and that the RP labels used by DSPN provide a strong signal for ACSA.

\begin{table}[h!]
\centering
\small
\begin{tabular}{lccccc}
\toprule
\textbf{Model} & \textbf{Supervision} & \textbf{Rest-14} & \textbf{Rest-15} &
\textbf{Rest-16} & \textbf{MAMS}\\
 \midrule
    ACSA-Generation & Supervised &78.43 & 71.91 & 73.76 & 70.30 \\
    JASen & Weakly supervised &26.62 & 19.44 & 23.23 & 14.74 \\
    AX-MABSA & Weakly supervised &49.68 & 42.74 & 36.47 & 29.74 \\
    DSPN (Ours) & \makecell{Unsupervised (Module 1) \\ Distantly Supervised
    (Module 2)} & 30.01 & 18.23 & 24.01 & 12.79 \\
    \bottomrule
\end{tabular}
\caption{\label{tab:noRP} ACSA accuracy results on datasets with no RP labels. Benchmark
results are from~\cite{kamila2022ax}.}
\end{table}


\subsection{Investigating Dimensional Bias}
\label{ssec:dimensionalBias}
DSPN uses star rating labels for training; however, the user rating may not be
consistent with the overall sentiment of the review text.
This is because the user may not have written all the
aspects in the review, or the user's sentiment is heavily dominated by a certain
aspect~\citep{ge2015measure}, 
thus generating noise when using distant supervision of RP labels for ACSA. 
Therefore, we conduct a simple additional
experiment. In the experiment, we utilize several unsupervised sentiment
analysis tools (VADER \citep{hutto2014vader}, TextBlob \citep{loria2018textblob},
and Zero-shot text classification \citep{yin2019benchmarking}) to directly
generate sentiment labels, which will replace the star rating labels given by
users for training.
We name the version of DSPN as UPN (U for unsupervised), and we report the ACSA
results of DSPN and UPN on TripDMS. Results in Table \ref{tab:fullyUnsupervised}
are mixed, demonstrating that while dimensional bias exists due to the inherent
limitations of star rating labels, its impact on DSPN’s overall performance is
minimal.

\begin{table}[h!]
	\centering
	\begin{tabular}{lcc}
		\toprule
		\bf Model&	\bf Label Source &	\bf Performance\\
		\midrule 
		DSPN	&Star ratings	& 50.0\\
		UPN	&TextBlob	&50.2\\
 		UPN	&VADER	&51.1\\
		UPN	&Zero-shot&	53.3\\
		\bottomrule
	\end{tabular}
	\caption{ACSA accuracy results when comparing DSPN to a fully unsupervised pyramid network (UPN). The DSPN result is identical to the result reported in our main results (Tables \ref{tab:highperformance} and \ref{tab:highefficiency}).}
	\label{tab:fullyUnsupervised}
\end{table}

\subsection{Module 1 Analysis}
\label{sec:found}

In this section, we investigate the choice of BERT as the foundation language model for 
our ACD component (Module 1).
Specifically, we compare the performance of our baseline DSPN pipeline with alternate 
versions where we replace BERT with two other language models: 
RoBERTa \citep{Liu2019RoBERTaAR} and ALBERT \citep{lanalbert}. 
Our results show that across both datasets, ACD performance is 
consistent (Table \ref{tab:found}).
For TripDMS, the larger RoBERTa model shows improved ACSA performance over BERT.
On the other hand, BERT outperforms RoBERTa for RP on ASAP and TripDMS. 
While ALBERT performs well on ACD, performance drops for ACSA and RP on both datasets. 
This result demonstrates that the choice of language model for Module 1 in  DSPN is part of the overall design consideration.
DSPN's modularity means that it can be adapted to account for improvements in contextual embedding models to further improve pipeline performance across tasks.

\begin{table}[h!]
	\centering
	\begin{tabular}{ccccccc}
		\toprule
		\bf \multirow{2}{*}{Dataset} &	\bf  \multirow{2}{*}{Foundation} &	\bf ACD  & \bf ACSA & \bf RP & \bf Training Time & \bf Params \\
        ~ & ~ & (F1) & (Acc) & (Acc) & (Minutes) & (M) \\
		\midrule 
		\multirow{3}{*}{TripDMS}	& BERT	& 92.7 & 50.0 & 71.7 & 38.7 & 109.5\\
		~ & RoBERTa & 92.7 & 51.6 & 71.6 & 38.8 & 124.7 \\
            ~ & ALBERT & 92.7 & 49.8 & 70.6 & 39.3 & 11.7 \\
            \midrule
            \multirow{3}{*}{ASAP} & BERT & 79.4 & 64.6 & 80.8 & 60.5 & 102.3\\
		~	& RoBERTa & 79.4 & 64.0 & 80.7 & 60.5 & 106.5\\
            ~ & ALBERT & 79.4 & 62.3 & 78.7 & 56.1 & 10.56\\
		\bottomrule
	\end{tabular}
	\caption{Results of different foundations. ACD, ACSA, and RP refers to aspect-category detection, aspect-category sentiment analysis, and rating prediction respectively.}
	\label{tab:found}
\end{table}

\section{Conclusion and Future Work}
\label{sec:conclusion}

In this paper, we introduce unified sentiment analysis to integrate
three key tasks within sentiment analysis. To model unified sentiment analysis, we propose a
Distantly Supervised Pyramid Network (DSPN) that shows an efficiency advantage
by only using star rating labels for training. Experiments conducted on two
multi-aspect datasets demonstrate the effectiveness of the pyramid network and
the strong performance of DSPN on RP and ACD
with only RP labels as supervision. DSPN's
performance demonstrates the validity of considering sentiment analysis
holistically, and this empirical evidence shows that it is possible to use a
signal from a single task (RP) to learn three distinct tasks effectively and
efficiently. We hope this paper spurs more research on novel unified sentiment analysis approaches
as well as the design of novel tasks that leverage one label source for
efficient learning for multiple tasks.

We offer several notable substantive implications for practitioners. First, our
framework does not require extensive and costly manual aspect-level annotations and can help
managers efficiently generate actionable, granular insights from customer
feedback. This is particularly valuable for organizations aiming to enhance
customer experiences without significant investments in manual data labeling and
organizations that do not have a sizable budget for market research. Second, our
framework informs the application of multi-task learning to integrate related
tasks into a unified framework, which can be applied to other domains where data
interdependencies exist. For example, an AI tool assisting with hiring could
leverage labeled final outcomes (e.g., hired or not) in a pyramid structure to
better model aspects such as cover letter appropriateness, relevant keyword
extraction from CVs, or appropriate candidate recognition for advertised
positions. In these applications, a pyramid structure would need to be augmented
with earlier labels. Still, the signal from the top of the pyramid (i.e., hired
or not) could guide labeling decisions, mirroring the efficiency and
interpretability achieved in sentiment analysis while addressing domain-specific
challenges.


This paper also has several limitations that shed light on promising avenues for
future research. One limitation relates to data availability. While there are a number
of datasets for ACSA and datasets for RP separately in the literature, 
datasets rarely support unified sentiment analysis. 
In fact, TripDMS and ASAP are the only datasets we could find---after a thorough review of the literature---that included labels for both aspect- and document-level sentiment. 
Therefore, we were restricted to TripDMS and ASAP as the only two datasets
available for our main evaluation.
Recent research has called for more work developing aspect-level datasets \citep{chebolu2023review}; we encourage the research community to develop datasets with labels at both levels for a more robust resource for training a variety of tasks and
to drive further research in unified sentiment analysis model development and evaluation. 
Next, DSPN requires predefined aspects, limiting its ability to discover novel aspects in new domains.
This assumption aligns with practical scenarios where domain experts can provide initial guidance, but may hinder applications requiring fully unsupervised discovery. 
Future work could consider combining automated aspect detection with human validation to address this challenge. 
What's more, with advancements in large language models and generative AI, 
future research could investigate how combining generative AIs with DSPN can leverage their strengths while mitigating these limitations, ultimately advancing sentiment analysis tasks~\citep{bhargava2025exploring}.
Finally, while the pyramid structure of DSPN allows for distantly-supervised modeling of ACSA, as designed the model cannot account for supervision for ACSA, even if some labels are available. 
Future work looking to incorporate weakly supervised techniques for ACSA into DSPN can benefit both from the unified modeling of sentiment and also from the signal provided by weakly supervised ACSA methods.  

\section*{Acknowledgements}
The authors would like to thank Lei Wang, Qiangming Yan, and Shuang Zheng for
their helpful comments on earlier versions of this work.


\bibliographystyle{ACM-Reference-Format}
\bibliography{references}

\appendix
\section{Additional Benchmarking}
\label{sec:ab}
Tables \ref{tab:rp}, \ref{tab:acsa}, and \ref{tab:acd} present the comprehensive
results of our benchmarking. We selected our pipeline models from these
benchmarks based on predictive performance and efficiency.

\begin{table*}[h!]
	\centering
	\small 
    \begin{tabular}{lcccccc}
		\toprule 
		         &          & \bf TripDMS &            &          & \bf ASAP   &
		         \\
        \cmidrule(lr){2-4} \cmidrule(lr){5-7} \bf Model               & \bf
		Accuracy & \bf Params  & \bf Train Time & \bf Accuracy & \bf Params &
		\bf Train Time  \\
		\midrule 
		 DSPN          & 71.7     & 109.5M   & 38.7min      & 80.8     & 102.3M   &
		 60.5min       \\
		 BERT-Feat     & 70.0     & 109.5M  & 3.5min      & 78.6     & 102.3M  &
		 2.1min       \\
		 BERT-FiT      & 72.2     & 109.5M     & 4.1min      & 80.7       & 102.3M &
		 6.4min       \\
		 BERT-ITPT-FiT & 72.9     & 112.7M   & 35.3min     & 81.6     & 103.5M    &
		 38.4min      \\
		 \bottomrule
	\end{tabular}
	\caption{Comprehensive RP results}
	\label{tab:rp}
\end{table*}


\begin{table*}[h!]
	\centering
	\small 
	\begin{tabular}{lcccccc}
		\toprule 
		         &          & \bf TripDMS &            &          & \bf ASAP
		         &             \\
        \cmidrule(lr){2-4} \cmidrule(lr){5-7} \bf Model                 & \bf
		Accuracy & \bf Params  & \bf Train Time & \bf Accuracy & \bf Params  &
		\bf Train Time  \\
		\midrule 
		DSPN            & 50.0     & 109.5M   & 38.7min      & 64.6     & 102.3M    &
		60.5min       \\
		End2end-LSTM    & 54.7     &  3.0M   & 0.2min       & 58.5     & 4.3M   &
		0.3min        \\
		End2end-CNN     & 43.4     & 3.62M   & 0.2min       & 45.2     & 4.9M   &
		0.3min        \\
		GCAE            & 58.0     & 15.3M   & 7.6min       & 74.5     & 15.3M    &
		6.1min        \\
		ACSA-Generation & 64.2     & 225.3M    & 59.5min     & 75.9     & 225.3M &
		51.3min     \\
		\bottomrule
	\end{tabular}
	\caption{Comprehensive ACSA results}
		\label{tab:acsa}
\end{table*}


\begin{table*}[h!]
	\centering
	\small 
	\begin{tabular}{lcccccc}
		\toprule
		     &      & \bf TripDMS &            &      & \bf ASAP   &
		     \\
        \cmidrule(lr){2-4} \cmidrule(lr){5-7} \bf Model           & \bf F1   &
		\bf Params  & \bf Train Time & \bf F1   & \bf Params & \bf Train Time
		\\
		\midrule 
		DSPN & 92.7 & 109.5M & 38.7min      & 79.4 & 102.3M   & 60.5min  \\
		ABAE      & 91.2 & 14.8M    & 36.2min      & 79.4 & 14.8M   & 37.4min       \\
		ABAE-BERT & 92.7 & 109.5M   & 35.2min      & 79.4 & 102.3M  & 54.5min      \\
		\bottomrule
	\end{tabular}
	\caption{Comprehensive ACD results}
		\label{tab:acd}
\end{table*}


\section{Word2vec Version of DSPN and ABAE}
\label{sec:dspnw2v}

Module 1 in DSPN implements an enhanced version of ABAE that replaces Word2vec with BERT as the embedding generator and also restricts the aspect matrix dimensions to the number of known aspects in the dataset. 
Further, it relies on keywords to initialize the aspect embeddings.
However, it may be the case that for a new dataset, the number of aspects is not known.
In this section, we compare an unsupervised version of DSPN's Module 1 to ABAE to demonstrate improved semantic coherence by including distant supervision via RP labels.
To do so, we replace BERT in Module 1 with Word2vec, so that we can compare learned aspect embeddings with static word embeddings to identify representative words for each topic~\citep{he2017unsupervised}.

Table \ref{tab:dspnw2v_coherence} presents the performance comparison between Word2vec-based DSPN and ABAE in terms of semantic coherence~\citep{mimno2011optimizing,rosner2014evaluating}.
Coherence is an automated metric of topic quality based on word co-occurrences that has been shown to correlate with human judgment. 
Specifically, coherence is defined as:

\begin{equation}
    C(T) = \sum_{m=2}^M \sum_{l=1}^{m-1} \log \frac{D(w_m, w_l) + \beta}{D(w_l)}
\end{equation}

\noindent
where $T$ is an ordered list of words for a topic, $D(w_m, w_l)$ and $D(w_l)$ are the co-document and document frequency counts, respectively, and $\beta$ is a smoothing parameter (e.g., $1$ or $\frac{1}{D}$).
Here, higher values indicate that the extracted aspects have better interpretability.
The more coherent the aspects are, the easier it is to explain the aspect. 
Taking the ``Room'' aspect in TripDMS as an example (Table \ref{tab:semantic_coherence}), the top words in the DSPN aspect appear to be more coherent. 
Therefore, we believe that incorporating distant supervision leads to more interpretable aspects.


\begin{table}[h!]
	\centering
	\begin{tabular}{lcc}
		\toprule
		\bf Dataset&	\bf  Model &	\bf Coherence Score\\
		\midrule 
		\multirow{2}{*}{TripDMS}	& DSPN	& \textbf{-643.3}\\
		~	& ABAE	& -651.4\\
        \midrule
 		\multirow{2}{*}{ASAP}	& DSPN	& \textbf{-740.2}\\
		~	& ABAE & -741.5\\
		\bottomrule
	\end{tabular}
	\caption{Coherence scores of DSPN and ABAE with Word2vec as foundaion.}
	\label{tab:dspnw2v_coherence}
\end{table}

\begin{table}[h!]
	\centering
	\begin{tabular}{lcc}
		\toprule
		\bf Aspect &	\bf  Model &	\bf Words \\
		\midrule 
		\multirow{2}{*}{Room}	& DSPN	& gloomy, damp, veneer, tatty, mildew, furniture, wood\\
		~	& ABAE	& threadbare, expose, grey, sewage, badly, mildew, odour\\
		\bottomrule
	\end{tabular}
	\caption{Illustration of semantic coherence.}
	\label{tab:semantic_coherence}
\end{table}

\end{document}